\documentclass[10pt,twocolumn,letterpaper]{article}

\usepackage[pagenumbers]{cvpr}

\usepackage{caption}
\usepackage{subcaption}
\usepackage{graphicx}
\usepackage{amsmath}
\usepackage{amssymb}
\usepackage{booktabs}
\usepackage{enumitem}
\usepackage{duckuments}
\usepackage{adjustbox}
\usepackage{algorithm}
\usepackage{makecell}

\usepackage[table]{xcolor}
\definecolor{cvprblue}{rgb}{0.21,0.49,0.74}
\usepackage[pagebackref, breaklinks=true, colorlinks, citecolor=citecolor, linkcolor=linkcolor, bookmarks=false]{hyperref}
\definecolor{citecolor}{HTML}{0071BC}
\definecolor{linkcolor}{HTML}{ED1C24}

\def\redc{\bf\cellcolor[HTML]{FF999A}}
\def\orangec{\it\cellcolor[HTML]{FFCC99}}
\def\yellowc{\cellcolor[HTML]{FFF8AD}}
\definecolor{detcolor}{gray}{.9}
\newcommand{\diffcell}[1]{\cellcolor{detcolor}{#1}}

\newcommand{\tablestyle}[2]{\setlength{\tabcolsep}{#1}\renewcommand{\arraystretch}{#2}\centering\footnotesize}
\newlength\savewidth\newcommand\shline{\noalign{\global\savewidth\arrayrulewidth
  \global\arrayrulewidth 1pt}\hline\noalign{\global\arrayrulewidth\savewidth}}

\usepackage{etoolbox}
\makeatletter
\AfterEndEnvironment{algorithm}{\let\@algcomment\relax}
\AtEndEnvironment{algorithm}{\kern2pt\hrule\relax\vskip3pt\@algcomment}
\let\@algcomment\relax
\newcommand\algcomment[1]{\def\@algcomment{\footnotesize#1}}
\renewcommand\fs@ruled{\def\@fs@cfont{\bfseries}\let\@fs@capt\floatc@ruled
  \def\@fs@pre{\hrule height.8pt depth0pt \kern2pt}%
  \def\@fs@post{}%
  \def\@fs@mid{\kern2pt\hrule\kern2pt}%
  \let\@fs@iftopcapt\iftrue}
\makeatother

\usepackage[stretch=10,shrink=10,step=1]{microtype}

\microtypesetup{
  babel=true,
  protrusion=true,
  expansion=true,
  tracking=true,
  spacing=true,
  factor=1200,
  final
}

\SetTracking{encoding={*}, shape=sc}{40}

\SetExtraSpacing[unit=space]{encoding=*}{
  \spaceshrink=30,
  \spacestretch=10
}

\title{FlowDet: Unifying Object Detection and Generative Transport Flows}

\author{Enis Baty\\
CVSSP, University of Surrey\\
{\tt\small e.baty@surrey.ac.uk}
\and
C. P. Bridges\\
SSC, University of Surrey\\
{\tt\small c.p.bridges@surrey.ac.uk}
\and
Simon Hadfield\\
CVSSP, University of Surrey\\
{\tt\small s.hadfield@surrey.ac.uk}
}

\begin{document}
\twocolumn[{
\maketitle
\vspace{-9mm}
\begin{figure}[H]
\hsize=\textwidth
\centering
\begin{subfigure}{0.3\textwidth}
    \centering
    \includegraphics[width=1.00\textwidth]{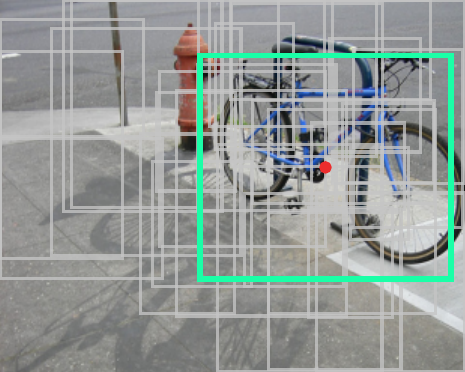}
    \caption{Non-Generative Detectors}
    \label{fig:1a}
\end{subfigure}
\hspace{0.1mm}
\begin{subfigure}{0.3\textwidth}
     \centering
     \includegraphics[width=1.00\textwidth]{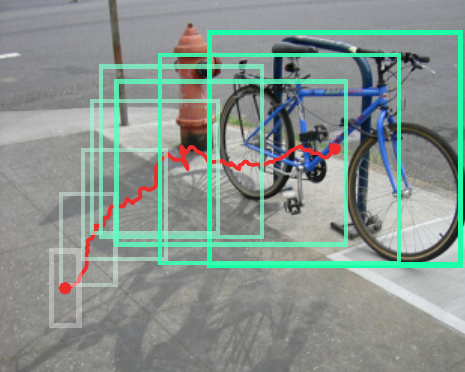}
     \caption{DiffusionDet}
     \label{fig:1b}
\end{subfigure}
\hspace{0.1mm}
\begin{subfigure}{0.3\textwidth}
     \centering
     \includegraphics[width=1.00\textwidth]{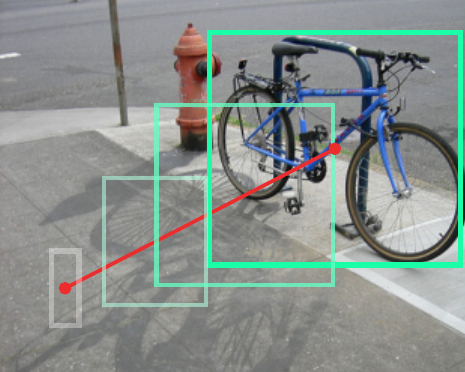}
     \caption{FlowDet}
     \label{fig:1d}
\end{subfigure}
\caption{Object detection paradigms:
(a) Traditional detectors are limited at inference time, predicting a single set
of boxes without the ability to dynamically iterate or transport distant boxes to better predictions.
(b) DiffusionDet \cite{diffdet} re-framed detection as a denoising process, enabling generative transport
at inference time, at the cost of complex stochastic paths.
(c) FlowDet generalises this generative formulation to Conditional Flow Matching,
providing shorter, straighter transport paths with greater performance using fewer steps.
}
\label{fig:teaser}
\end{figure}
\vspace{-1mm}
}]

\begin{center}
\section*{Abstract}
\end{center}
\vspace{-2mm}
{\it
We present FlowDet, the first formulation of object detection using modern Conditional Flow
Matching techniques. This work follows from DiffusionDet, which originally framed detection
as a generative denoising problem in the bounding box space via diffusion. We revisit and
generalise this formulation to a broader class of generative transport problems, while maintaining
the ability to vary the number of boxes and inference steps without re-training.
In contrast to the curved stochastic transport paths induced by diffusion, FlowDet learns simpler
and straighter paths resulting in faster scaling of detection performance as the number of
inference steps grows.
We find that this reformulation enables us to outperform diffusion based detection systems (as well
as non-generative baselines) across a wide range of experiments, including various precision/recall
operating points using multiple feature backbones and datasets. In particular, when evaluating under recall-constrained settings, we can highlight the effects of the generative transport without over-compensating with large numbers of proposals. This provides gains of up to +3.6\% AP and +4.2\% AP$_{rare}$ over DiffusionDet on the COCO
and LVIS datasets, respectively.
}    
\vspace{-2mm}
\section{Introduction}
\label{sec:intro}
\vspace{-1mm}
Modern works cast detection as a set prediction problem where a given set of proposals or
learnt embeddings are refined through multi stage heads to an output set of predictions. 
This enables end-to-end training of the detection model \cite{sparse-rcnn, detr},
in contrast to classical two-stage approaches \cite{fast-rcnn, faster-rcnn},
significantly improving detection quality. However, at inference time, the performance
of these detectors are fixed to the representational budget embedded at train time.

Recently, DiffusionDet \cite{diffdet} proposed an entirely new framing of the detection
problem. By casting detection as generative transport from a simple prior, to the data
distribution of target boxes, compute can be adjusted dynamically at test time, which has recently gained
significant traction in other deep learning domains \cite{wang2023selfconsistency, snell2024scalingllmtesttimecompute}.
Taking more refinement steps or drawing more samples can improve task performance without retraining, 
at the cost of compute. This also provides flexibility for a given model to be used in either limited
compute scenarios (edge inference, mobile devices) or unrestricted for upmost accuracy \cite{Han2021DynamicNN}.
Additionally, the generative transport framing removes the need for hand‑crafted proposals or
separately trained proposal region generation. As such, a simple prior distribution to sample proposals from
can be defined and the model learns the transformation to a target set of ground truth instances.

While diffusion has proven to be a strong generative framework, many works seek to reduce the computational
burden at inference time \cite{dpm-solver, dpm-solver-pp, prog-distill-diff, consistency-models}. For example,
the use of DDIM samplers over the canonical DDPM  reduce the number of steps by about 10-50$\times$ to produce
comparable output quality at inference time. \cite{ddim}
Although significant efforts have been made to improve upon diffusion, contemporary literature has
shifted to focus on the very formulation of diffusion to root out transport inefficiency. As such,
Conditional Flow Matching (CFM) \cite{cfm} introduces a generalisation of the diffusion process that
further improves the compute-accuracy trade off at integration time.

Conditional Flow Matching is a simulation‑free generative modelling framework that trains a time‑varying
velocity field to move samples from a simple prior to the data distribution. While diffusion samplers
trace curved, schedule‑induced trajectories, CFM allows for near‑linear paths by construction through
linear interpolations between prior and target samples as input interpolants during training.
Furthermore, instead of random source–target pairings that yield misaligned transport and meandering
flows, simple data-aligned priors and transport cost-aware path construction can improve sample efficiency
and accuracy at a fixed level of compute. Contrasting diffusion, off-the-shelf ODE solvers provide another
angle for precise accuracy-latency control at inference time without schedule tuning.

This work introduces FlowDet, a CFM-based detector generalising the definition of object detection as generative transport in
bounding box space. During training, we sample proposal boxes from a simple normal Gaussian distribution and
form linear interpolants to ground‑truth boxes at randomly sampled timesteps. These interpolants are
then used to train the detection head or decoder of our model then to predict the constant velocity flow
associated with each interpolant that transports a given bounding box sample from the prior distribution to
the target (i.e. ground truth) distribution. At inference time, an arbitrarily sized set of samples are
drawn from the prior distribution as proposals. These prior samples are then integrated through the latent
flow field embedded within the detection decoder to produce a set of clean bounding box and class label
predictions for a given image as shown in Figure \ref{fig:teaser}.

By adopting a generative approach, FlowDet becomes extremely flexible at inference time. Without re-training
or fine-tuning, the model can trade-off runtime against precision or recall, by adapting the number of inference
steps and source-boxes respectively. With the same ResNet50 \cite{resnet} backbone we surpass the performance of many strong
baselines in detection on the COCO \cite{mscoco} dataset. Specifically compared to DiffusionDet, we achieve
+0.1\%AP with identical configurations of evaluation boxes and steps and scaling further to a +0.4\%AP gain
with a higher number of steps.

We summarise our contributions as follows:
\begin{itemize}
    \item To the best of our knowledge, this work is the first to formally frame 
          detection as a generative transport task using Conditional Flow Matching, 
          generalising the box denoising formulation proposed by DiffusionDet.
    \item We ablate a broad range of possible data driven priors and matching mechanisms for
          coupling sources to targets, providing a solid basis for future generative detection
          approaches to build on.
    \item We evaluate on standardised benchmarks and backbones, demonstrating FlowDet's inference time flexibility by operating at varous precision/recall tradeoffs without model retraining. Specifically we demonstrate
          trading compute for recall and accuracy by varying proposal counts and integration steps respectively
          with a single checkpoint, and achieve notably higher transport efficiency in recall-constrained,
          few-proposal regimes.

\end{itemize}

\vspace{-0mm}
\section{Related Work}
\label{sec:rel-work}
\vspace{-0mm}
\subsection{Object Detection}
Classically, deep object detection approaches formed a two‑stage process that relied on classifying
hand-crafted region proposals through Selective Search \cite{selSearch, rcnn, fast-rcnn}
before rapidly progressing to Region Proposal Networks \cite{faster-rcnn}. These two-stage
detectors perform well with Region Of Interest (RoI) heads providing strong localisation and
robustness at the cost of greater compute through the two distinct proposal and refine stages.

Single-stage ``dense'' methods adopted direct prediction of boxes from multi-scale feature maps through heuristic
anchor grids. These evolved into anchor-free methods treating each location as an object centre,
achieving simpler, faster pipelines with competitive accuracy \cite{ssd, retinanet, yolo, fcos, centernet}.

The rise of the transformer architecture \cite{attn-is-all-you-need} has spawned many works that treat detection as a
set prediction problem leading to the development of DETR \cite{detr}. Through introducing box-to-box
interactions, DETR-family models significantly reduce redundancy within the prediction set and
exploit correlations between related objects. Though initially suffering from poor convergence
speed and high computational requirements, significant efforts improved both the architectural limitations, 
\cite{deformableDETR, efficientDETR, sparseDETR, sparse-rcnn, conditionalDETR, dynamicDETR, anchorDETR, dabDETR}
and set prediction supervision methods \cite{dnDETR, co-DETR, groupDETR, alignDETR, dyn-spa-rcnn}.

However, all prior works that rely on dense prediction depend on hand-crafted post-processing such as
Non-Maximal Supression (NMS). Fixed end-to-end methods are also constrained by rigid training time
hyper-parameter choices and their behaviour cannot be modulated during deployment.
Drawing parallels to image generation, DiffusionDet \cite{diffdet},
formulates detection as a denoising diffusion process \cite{ddpm} that iteratively refines noisy boxes to
object proposals. In this paper we exploit the same generative formulation for dynamic inference.
However, we generalise from basic diffusion denoising to CFM, allowing us to achieve competitive
results with very few inference steps.

\subsection{Diffusion}
Diffusion Denoising Probabilistic Models (DDPM) \cite{ddpm} frame generation as learning to
reverse a fixed Gaussian noising process, inspired by non-equilibrium thermodynamics \cite{sohldickstein15}.
Models are trained with a noise-prediction objective for noise that gradually corrupts ground truth samples
through a Markovian chain in the forward process. The reverse process then starts with pure noise and iteratively
``denoises'' to produce an output sample that can be influenced with conditioning to guide generation towards
specific instance types present in the original training set. 

Denoising Diffusion Implicit Models (DDIM) introduce a non‑Markovian, deterministic sampler that preserves
the original diffusion marginals while enabling few‑step generation and exact image-latent inversion 
providing a valuable speed-quality trade-off \cite{ddim}.
Subsequent work refines schedules, parameterizations, and solvers (e.g., improved diffusion, EDM), but most
contemporary systems still build on the DDPM objective with DDIM‑style fast samplers for efficient inference
and controllable guidance \cite{impr-ddpm,karras2022edm}.

Although DDIM enables relatively fast sampling, the efficiency of the diffusion process is still the leading
challenge in the field. Given the stochastic formulation of the task, the denoising process is computationally
expensive and provides poor results with few inference time sampling steps.

\subsection{Conditional Flow Matching} \label{sec:lit-cfm}
Continuous Normalising Flows (CNFs) model generation as the integration of a learned, time‑varying flow that
transports a simple prior to the data distribution \cite{chen2018neuralnode,grathwohl2018scalable,papamakarios2021flows}. 
Conditional Flow Matching (CFM), provides a simulation‑free objective to train CNFs
by regressing the velocity field along a user‑specified probability path. This unifies flow‑based and diffusion‑style
training within a single framework, where diffusion arises as a specific path choice, making diffusion a special case
of flow matching \cite{cfm,albergo2023building,albergo2023stochastic,gao2025diffusionmeetsflow}.
This perspective exposes explicit design levers such as prior distributions, paths, coupling strategies and supports
deterministic sampling via standard ODE solvers that allow for an accuracy/compute trade-off chosen at inference time.

Recent work refines the framework in three key areas: path design, coupling, and straight‑trajectory training.
Optimal‑Transport (OT) couplings and mini-batch OT improve alignment between source and target, shortening transport
and stabilizing training \cite{tong2024improving, wcfm}. Straight‑path objectives such as Optimal Flow Matching
and Rectified Flow reduce curvature and path length, improving sample efficiency and enabling few‑step or even
one‑step generation without sacrificing fidelity \cite{huang2025onestep,liu2023flow}. Conditional variants broaden
applicability, with Conditional Variable Flow Matching (CVFM) generalising to continuous conditioning with
unpaired data and reflected/variational rectified flows further tighten the train–test gap to improve solver robustness
\cite{cvfm,rfm24,guo2025variational}.

To the best of our knowledge, this work is the first to generalise the generative detection problem
to the broader category of flow-based methods. This enables us to adopt Conditional Flow Matching techniques, preserving
the test‑time flexibility established by diffusion‑based detectors
while learning more compact flow fields that require less recursive calls to traverse at inference time.

\vspace{-0mm}
\section{Methodology}
\label{sec:methodology}
\vspace{-0mm}
\subsection{Preliminaries} \label{sec:method-prelim}
Conditional Flow Matching (CFM) builds upon the formulation of Continuous Normalizing Flow (CNF),
which transform a probability density $p(x)$ via an ordinary differential equation
(ODE) $\frac{dx}{dt} = v_\theta(x, t)$ from a simple prior $p_0(x)$ at $t=0$ to the data distribution $p_1(x)$ at $t=1$ \cite{chen2018neuralnode}.
In both CNFs and CFM, the time-dependent vector field $v_\theta(x, t)$ is parametrised by a neural network, however
the key differences lie in the training of $v_\theta$ enabling the scaling of CFM for use in non-trivial settings.

CNFs enable flexible transformations but require computing the log-determinant (trace) of the Jacobian
$\frac{\partial v_\theta}{\partial x}$ to track probability densities via the instantaneous change-of-variables formula.
This is computationally expensive and typically approximated by stochastic trace estimation,
introducing training instability and scalability issues \cite{grathwohl2018scalable,papamakarios2021flows}.

CFM circumvents this intractability by training a model to regress the underlying vector field $v_\theta(x, t)$,
rather than modelling transport directly. The conditional probability paths $p_t(x \mid x_1)$ 
are trained 
between samples $x_0 \sim p_0$ and data points $x_1 \sim p_1$, with the regression objective
\begin{equation} \label{eqn:cfm-obj}
     \mathcal{L}_{CFM} = \displaystyle{E}_{t, x_1, x_0} \|v_\theta(x_t, t) - u_t(x_t \mid x_1)\|^2.
\end{equation}
This in turn requires no simulation, no explicit density evaluation, and no Jacobian computation \cite{cfm}.
A critical finding of prior CFM works is that we do not need to explicitly compute the optimal transport flow field
$u^*$ to serve as the ``ground truth'' target for our regression. Instead, we recover an equivalent flow field by
marginalising over random draws from $p_0$ and $p_1$ with a simple and tractable pairwise path.

Diffusion can be viewed as a special case of CFM under a Gaussian linear path \cite{gao2025diffusionmeetsflow}:
\begin{equation}
v(x_t \mid x_0)=\mathcal{N}(x_t \mid a_t x_0,\; s_t^2 I),\quad a_t=\sqrt{\bar{\alpha}_t},\; s_t^2=1-\bar{\alpha}_t,
\end{equation}
where $s_t$ and $a_t$ represent the single step, and cumulative noising weight respectively, at $t$.
DDPMs are then trained with the denoising objective
\begin{equation}
\mathcal{L}_{DDPM} = \tfrac{1}{2}\,\|v_\theta(x_t,t)-x_0\|_2^2,
\end{equation}
and reverse-time updates (e.g., DDIM) at inference.
In CFM, we choose the linear-Gaussian conditional path
\begin{equation} \label{eqn:cfm-interp}
\vspace{-0mm}
x_t=(1-t)\,x_0 + t\,x_1 ,\quad x_0\sim\mathcal{N}(0,I),
\vspace{-0mm}
\end{equation}
whose conditional target flow field is the constant interpolant
\begin{equation}
\vspace{-0mm}
u_t(x_t \mid x_1)= x_1 - x_0.
\vspace{-0mm}
\label{eq:interpolant}
\end{equation}
The neural network $v_\theta(x, t)$ is then trained through the objective in Eqn. \ref{eqn:cfm-obj}.
Because the underlying pairwise paths are linear, they by definition represent the shortest
possible paths between samples. The intuition behind CFM is that marginalising over a collection
of shortest possible paths, gives a resulting flow field that is close to the ideal optimal
transport mapping. These simple and compact ODE flow fields can then be traversed efficiently
in a few Euler steps at inference times with
\begin{equation}
x_{t+\Delta}=x_t+\hat{u}_\theta(x_t, t)\,\Delta.
\end{equation}
This contrasts the complex SDE fields induced by diffusion, which must be traversed via many DDPM steps.

\subsubsection{Detection with CFM}

Object detection can be parameterized as conditional set prediction. Given an image I $\in \mathbb{R}^{H \times W \times 3}$
and its feature map $\mathbf{p}$, we define a probability distribution $p(\mathcal{Y}\mid F)$ over finite, order-agnostic sets
$\mathcal{Y} = \{(b_i, c_i)\}_{i=1}^{M}$ with M $\in \mathbb{N}_0$ varying per image.
Each element comprises a bounding box $b_i = (c_x, c_y, w, h) \in \mathbb{R}^4$ and a class label $c_i \in \{1,\dots,K\}$,
with $(c_x, c_y)$ defining the center coordinates and $(w, h)$ the width/height of the $i$-th bounding box within a particular
set.

In this work, to apply the theoretical framework of CFM to the task of object detection, we treat $x_1$ as a ground truth
set of bounding boxes $\mathcal{Y}$ and $x_0$ as a set of proposals sampled from a simple prior distribution.
Training a neural network through the regression
objective in Eqn.~\ref{eqn:cfm-obj} yields a flow field that transports samples from the prior to predicted endpoints $\hat{x}_1$.
To further align this formalization of CFM to the detection task, rather than predicting a velocity, we predict the
endpoint $\hat{x}_1$ and class label $\hat{c}$ while recovering the velocity algebraically for sampling during inference as
\begin{equation}
\hat{u}_t = \hat{x}_1 - x_0,\ \text{where }\hat{x}_1 = v_\theta(x_t, t),
\end{equation}
which enables standard detection-oriented set losses on $\hat{x}_1$,$\hat{c}$, tightly coupling the training objective to
the task \cite{detr,diffdet,hybridmatching2023}.
\vspace{-2mm}
\subsection{Architecture}
\begin{figure}[t]
  \centering
  \includegraphics[width=\linewidth, trim={0mm 1mm 2.5mm 0mm}, clip]{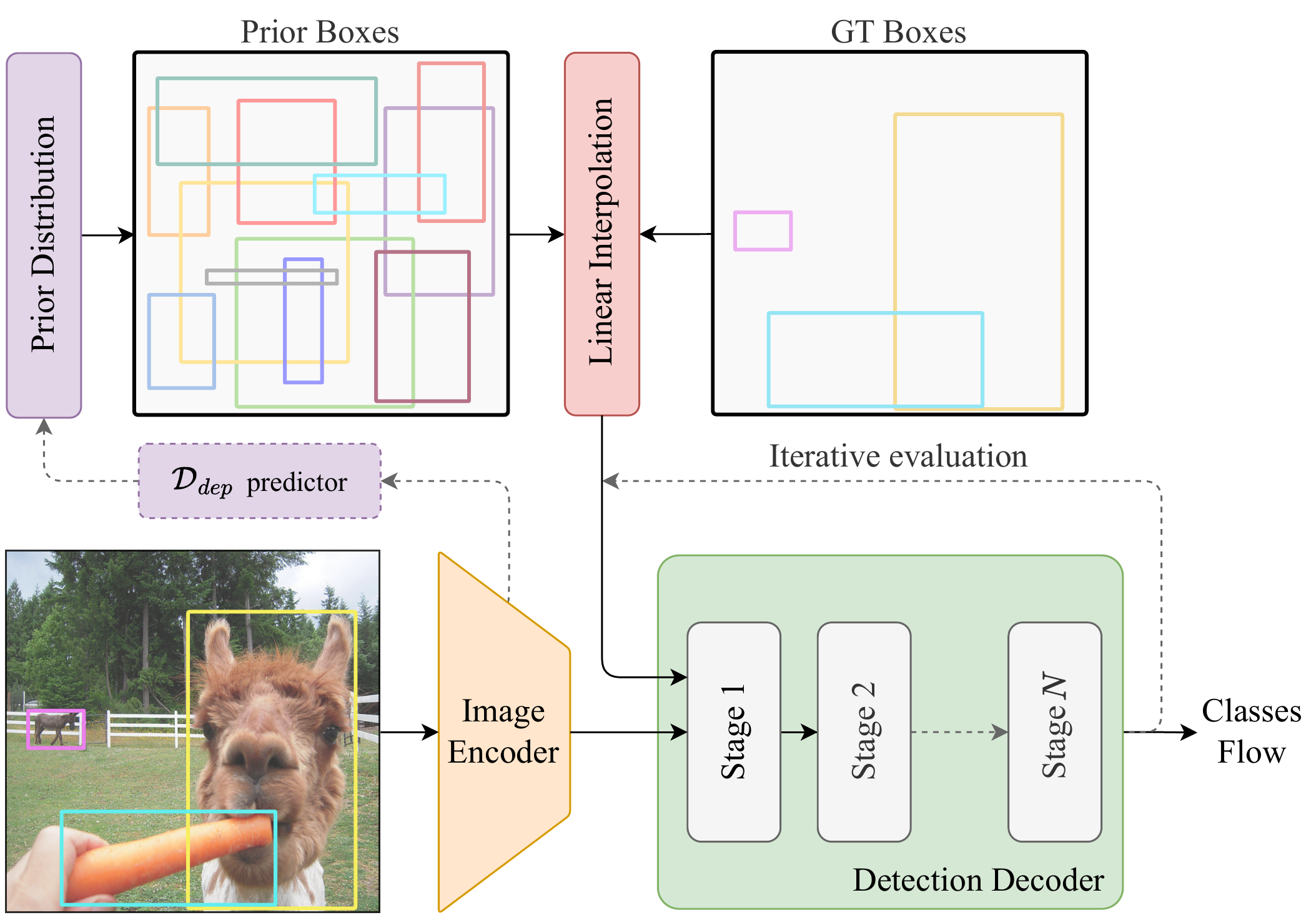}
  \caption{Architecture block diagram of FlowDet.}
  \label{fig:arch}
  \vspace{-4mm}
\end{figure}

To allow fair comparison to previous generative detectors like DiffusionDet, we adopt the Sparse R-CNN 
base architecture. As shown in Figure \ref{fig:arch}, the model can be viewed as two primary components,
an image encoder and detection decoder along with a prior statistics predictor in the case of data-dependant
priors.

\subsubsection{Image Encoder}

A standard vision backbone (ResNet \cite{resnet} with a Feature Pyramid Network \cite{fpn}) maps images
$\mathbf{I} \in \mathbb{R}^{C \times H \times W}$ to rich multi-scale feature maps 
\begin{equation}
    \{\,\mathbf{p}_{\ell}\,\}_{\ell \in \mathcal{L}} = \mathrm{FPN}\big(\textrm{Backbone}(\mathbf{I})\big),
\end{equation}
where $\ell$ indicates the featuremap at a particular pyramid level, and $\mathbf{p}_{\ell} \in
\mathbb{R}^{D \times H_{\ell} \times W_{\ell}}$ are time-invariant and reused across all timesteps.

\subsubsection{Prior Distribution} \label{sec:method-priors}

Unlike diffusion models, CFM allows for the exploration of alternate choices of prior distributions, including
data-derived or data-dependent distributions. Theoretically, through closer alignment of the prior
distribution to the ground truth distribution, the conditional paths between a sample point $x_0$ and
data point $x_1$ are both shorter and straighter. These simplified paths improve the efficiency
of the integration process, at the cost of reducing exploration and increasing redundancy between boxes.

In this section we will formulate a range of possible prior distributions, which will then be ablated in Section
\ref{sec:abl-priors}. Our base prior distribution is defined as $\mathcal{D}_{GaussN} = \mathcal{N}(0,1)$ inline
with diffusion. We define ``data-derived'' as a set of prior distributions that are pre-computed from the dataset
statistics and applied uniformly over all incoming samples. Meanwhile we define ``data-dependant'' as distribution
statistics that are learnt during training.

More specifically we implement the data-derived prior $\mathcal{D}_{der}$
as a Gaussian whose mean and standard deviation match those aggregated across all the ground-truth bounding boxes
in the dataset. We extend this idea with $\mathcal{D}_{der-sb}$ as a Mixture of Gaussians, each of which is fit to
a size-bucketed subset of the ground-truth bounding boxes.
In contrast, a data-dependant prior $\mathcal{D}_{dep}$ predicts the statistics per image
\begin{equation} \label{eqn:data-dep-prior}
    (\hat{\mu}, \hat{\sigma}) = \mathrm{Sigmoid}(\mathrm{MLP}(\mathrm{GAP}(\mathbf{p}_{-1}))),
\end{equation}
where GAP takes a Global Average Pool of the coarsest feature map $\mathbf{p}_{-1}$ to generate a feature vector
and MLP is a simple two layer Multi Layer Perceptron.
We outline the supervision of
$\mathcal{D}_{der-sb}$ during training in Section \ref{sec:training-supervision}.

To ensure that RoIs derived from bounding boxes crop valid regions of images, the prior distributions are sampled
via inverse Cumulative Distribution Function (CDF) sampling from a factorized truncated normal.

\subsubsection{Detection Decoder}

We use a DynamicHead architecture similar to Sparse R-CNN \cite{sparse-rcnn} which consists of six stages.
Stage $n$ of the head crops the feature maps from the image encoder, $\mathbf{p}_\ell$, to obtain RoI features 
$\mathbf{r}^n_t = \mathrm{RoIAlign}(\mathbf{p}_\ell, x^n_t)$. The RoI features are then processed into
object features $h^n_t$ through an RCNNHead \cite{sparse-rcnn} with multi-head shared attention, dynamic instance
interaction and a feedforward network,
\begin{equation}
    h^n_t = \mathrm{FFN}(\mathrm{DynamicHead}(\mathrm{MHSA}(\mathbf{r}^n_t, \tilde{h}^{n-1}_{t})).
\end{equation}

To produce $\tilde{h}^{n}_{t}$, we condition $h^n_t$ with the current timestep $t$ by first applying sinusoidal embedding \cite{attn-is-all-you-need}
$\tau = \phi(t)$ then, following FiLM \cite{FiLM}, we project the embedding to a scale $\gamma$ and shift $\beta$ such that
\begin{equation}
    \tilde{h}^n_t = \gamma(\tau) \odot h^n_t + \beta(\tau).
\end{equation}

The time-conditioned object features $\tilde{h}^n_t$ are then jointly used to predict a bounding box offset ($\Delta \hat{x}$) and classification label ($\hat{c}$) from the CFM model
\begin{equation}
\begin{aligned}
  (\Delta \hat{x}_t^{\,n}, \hat{c}_{\,t}^{\,n}) &= v^n_{\theta}\!\big(\tilde{h}^{n-1}_t).
\end{aligned}
\end{equation}
The output offsets are then applied to the bounding box proposals
before being iteratively fed to the next head stage
\begin{equation} \label{eqn:det-stage}
\begin{aligned}
  \hat{x}_t^{\,n} &= \hat{x}_t^{\,n-1} + \Delta \hat{x}_t^{\,n}.
\end{aligned}
\end{equation}
We train the model using a single randomly selected timestep per sample, to ensure broad coverage of the flow field.
At inference time, we draw $N_{eval}$ timesteps from a linear schedule ($[t_1,...,t_{N_{eval}}]$). The full head
(including all stages) is then called in turn, with the output of the final head stage feeding in as input to the
first head stage at the next time step. More formally, the initial conditions for Eqn.~\ref{eqn:det-stage} at each
timestep are $\hat{x}_{t_i}^{\,0} = \hat{x}_{t_{i-1}}^{\,N} + \Delta \hat{x}_{t_{i-1}}^{\,N}$.

\subsection{Training}

During training, the model is fed with the interpolants $u_t$ constructed according to the conditional paths
defined in Equation \ref{eq:interpolant}. As such, we replace the traditional Region Proposal
Network \cite{faster-rcnn} or learnt bounding box proposals \cite{sparse-rcnn} with a random set of source
boxes sampled from the chosen prior distribution.
Each source sample, $x_0$, from the prior distribution must be paired with a valid ground truth
sample, $x_1$ as outlined below.

\subsubsection{Ground Truth Padding}
In practice, each image contains a varying number of ground truth boxes, $N_{gt}$. Since we must
assign each source sample to a target box, we must pad the ground truth set to match the chosen
number of training proposals, $N_{train}$. We follow DiffusionDet's empirical findings and pad
$N_{gt}$ with random samples of the source distribution which proves more effective than crude
duplications of the ground truth boxes.

\subsubsection{Interpolant construction}
\label{sec:matching}
In the same way that CFM supports a range of prior distributions, it is possible to employ a range of matching strategies between source and target samples. In its vanilla form, CFM employs random pairing of $x_1$ samples against $x_0$ samples. We call this matching strategy $\mathcal{M}_{rand}$. However, this can be slow to converge as many training pairs lie far from the optimal transport path. To accelerate convergence we can implement mini-batch OT \cite{tong2024improving} by employing Hungarian matching between source and target sets. We refer to this matching strategy as $\mathcal{M}_{hung}$. This can focus training budget on the pairs that are closer to the optimal transport paths, but it can leave less common areas of the flow field underconstrained. When employing hungarian matching, we explore a range of different similarity measures between boxes including: centrepoint distance ($\mathcal{M}_{hung-c}$) which is scale agnostic, generalised IoU (gIoU) \cite{gIoU} + bounding box difference ($\mathcal{M}_{hung-g}$) and InterpIOU ($\mathcal{M}_{hung-i}$) \cite{InterpIoU}. Both gIoU and InterpIoU are intended to measure overlap, but in a manner that decays smoothly for non-overlapping boxes.

Regardless of the matching strategy, the paired samples are then constructed into CFM interpolants as defined in Eqn.~\ref{eqn:cfm-interp}
with $t \sim U[0,1]$ in contrast to the cosine schedule commonly adopted with diffusion.

\subsubsection{Training supervision} \label{sec:training-supervision}
While the Gaussian and data-derived prior distributions outlined in Sec. \ref{sec:method-priors} are static
and have no trainable parameters, the data-dependent prior is supervised in isolation to the
detection decoder as sampling the truncated Gaussian distributions prevents gradient flow.
Therefore, the predicted $\hat{\mu},\hat{\sigma}$ are regressed against the mean and std. dev. of ground truth boxes within the batch.

The outputs from each stage in the detection decoder ($\hat{x}_1, \hat{c}$)
are independently supervised against the ground truth set ($x_1, c$) through the standard set prediction
loss popularised by \cite{detr}.
The full definition of the training loss (including the one-to-many assignment of predictions to ground-truth) is provided
in Appendix \ref{app:train-loss}. However, in summary, it comprises 
\begin{equation} \label{eqn:set-cost}
    \mathcal{L}_{match} = \lambda_{cls} \cdot \mathcal{L}_{cls} +
                  \lambda_{\ell_1} \cdot \mathcal{L}_{\ell_1} +
                  \lambda_{gIoU} \cdot \mathcal{L}_{gIoU}.
\end{equation}
Here, we define $\mathcal{L}_{cls}$ as focal loss between $\hat{c}$ and $c$, $\mathcal{L}_{\ell_1}$ and
$\mathcal{L}_{gIoU}$ as the $\ell_1$ and gIoU loss, respectively, between $\hat{x}$ and $x$. $\lambda_{cls}$,
$\lambda_{\ell_1}$ and $\lambda_{gIoU}$ define their respective weights.

\subsection{Inference}
At inference, we sample an arbitrary sized set of $N_{eval}$ boxes from the prior distribution
and integrate them through the latent flow field within the detection decoder with an ODE solver.
Given the strong single step performance of the detection decoder, we implement a simple Euler
integrator and ablate higher order solvers in Sec. \ref{sec:abl-ot}.

During integration with multiple sampling steps, weak predictions are discarded
per step based on intermediate classification scores and the set of proposals is replenished
with new samples from the prior distribution following \cite{diffdet}. Outputs across each sampling
step are aggregated as an ensemble set of predictions with Non-Maximal Suppression (NMS) to reduce
trivial duplicates between steps. The final set of bounding boxes post-filtering are resized
and clipped such that the predictions are bounded to the image size.
 
\section{Experiments}
\label{sec:experiments}

We benchmark FlowDet on both standard (COCO) and fine-grained (LVIS) detection tasks.
We compare with common baseline detectors covering a range of performance intervals. We also provide detailed
comparisons against DiffusionDet to highlight the stronger accuracy/compute trade-off
within the scope of generative object detection. Following this, we explore and ablate design choices in Section \ref{sec:ablations}.

Common Objects in COntext (COCO) \cite{mscoco} has been established as the de facto benchmark for object
detection and other tasks (panoptic segmentation, keypoint detection, etc.). We follow the
commonplace practice of using the 2017 train/val split that contains 118K/5k images
respectively. Objects within images are classified into 80 categories with images including
both sparse and crowded context-rich scenes. Evaluation is quantified with the Mean Average Precision
(mAP, often denoted as AP) metric that reports AP averaged over IoU thresholds ranging
from 0.5:0.95 and additional small/medium/large scale stratified AP measures.

The Large Vocabulary Instance Segmentation (LVIS) \cite{lvis} dataset extends COCO's annotations to a
large-vocabulary, long-tail setting that further exemplifies the frequency imbalance of
objects in real-world imagery. In this case we follow the standard protocol and report frequency stratified AP measures
(AP$_{rare}$, AP$_{common}$, AP$_{frequent}$) that differentiate between performance on classes
with a high number of instances.

\vspace{-1mm}
\subsection{Experimental Setup}
\vspace{-1mm}
We evaluate our technique using two backbone models: ResNet50 and ResNet101 with FPN, all initialised with
pre-trained weights from ImageNet \cite{imagenet} pre-training as outlined in Detectron2's
model zoo \cite{det2}. All models are initialised with 6 head stages and are trained to a standard COCO
3x schedule, adjusted for a batch size of 8, resulting in approx. 532k iterations for a single GPU at a learning rate
of $2.5 \times 10^{-5}$ with weight decay of $1\times10^{-4}$. The learning rate is linearly warmed up
for 1000 iterations then scheduled to decay by a factor of $0.1\times$ after $2/3$ and $8/9$ of the
total training iterations. Following \cite{detr}, we set the matching cost and loss weightings
defined in Eqns. \ref{eqn:set-cost} to $\lambda_{cls}=2.0$, $\lambda_{\ell_1}=5.0$
and $\lambda_{gIoU}=2.0$. Data augmentations during training consist of random horizontal flips,
crops and scale jitter, that resizes images such that the shortest and longest edges are between
480-800 pixels and 1333 pixels respectively.

In terms of CFM specific setup, we chose to set our prior distribution to a standard Gaussian,
with random pairing, and a simple Euler ODE integrator for inference. These settings are
derived from our ablations conducted in each area, as described in Section \ref{sec:ablations}.

At inference, we evaluate the model with various numbers of boxes and integration steps to
highlight the flexibility of FlowDet without re-training, following \cite{diffdet}.
To clearly demonstrate the capabilities of generative detectors, we constrain recall and isolate
the effects of the generative transport mechanism 
from the effect of having many highly redundant proposals.
We therefore evaluate with
$N_{eval} \in \{50, 75, 100\}$ proposal boxes, while still training with $N_{train}=300$ 
to accelerate convergence and ensure flow field coverage.
The post processing pipeline includes default rescaling and clipping of predictions with
class-wise NMS, with an IoU threshold of 0.6, to suppress duplicates introduced by the top-k
matching in the set prediction losses. 

\begin{figure}[t]
  \centering
  \includegraphics[width=\linewidth, height=5cm, trim={2.5mm 2.5mm 5mm 2.5mm}, clip]{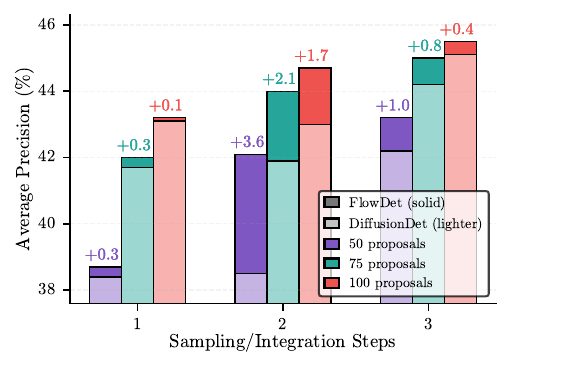}
  \vspace{-5mm}
  \caption{\textbf{Plot of AP against integration steps between FlowDet and DiffusionDet.} We plot
           AP for $N_{eval} \in \{50, 75, 100\}$.}
  \label{fig:ap-vs-step}
  \vspace{-5mm}
\end{figure}

  \vspace{-2mm}
\subsection{Results}
  \vspace{-1mm}
We first present our results on the validation split of COCO using 2 different feature backbones,
in Table \ref{tab:coco} alongside strong non-generative baselines and DiffusionDet as a generative
comparison. FlowDet surpasses all the non-generative baselines across almost all metrics and backbones.
For example, a margin of +0.5\% AP over Sparse R-CNN with three integration steps. Similarly, FlowDet
outperforms DiffusionDet across almost all metrics, at all operating points, with both feature backbones.
In the single-step base case (where the flow field is weakly utilised) we achieve a slight advantage of +0.1\% AP.

When introducing a few iterative evaluation steps, FlowDet scales more rapidly than DiffusionDet,
resulting in a +0.4\% AP gain in the 3 step setting. These strengths in iterative evaluation are repeated
with stronger backbones, validating the potential for FlowDet to scale with more expressive image encoders.
Figure \ref{fig:ap-vs-step} shows that in a recall‑constrained regime with only 50 proposals, FlowDet delivers
a +3.6\% AP gain over DiffusionDet at comparable compute, indicating more efficient usage of proposals with lower redundancy. When the proposal budget is low, the transport paths from initial boxes to targets are naturally longer,
and the detection head has fewer opportunities to ``shortcut'' the transport via lucky source samples. This highlights how
CFM’s more direct flows yield more reliable localization. As the number of proposals increases, both
methods benefit from denser shortcuts, but FlowDet maintains a consistent advantage across steps and proposal counts.

Evaluating on LVIS with like-for-like settings as \cite{diffdet}, FlowDet again shows
strong scaling performance, outperforming other generative detection models across almost all metrics and operating points.
Of particular note is FlowDet's excellent performance on rare item categories, with improvements as high as 4.2\% AP$_{rare}$ in some cases.

\begin{table}[t]
\tablestyle{3.7pt}{1.1}

\begin{adjustbox}{trim=0pt 0pt 0pt 0pt, clip} 
\begin{tabular}{@{}lcccccc@{}}
\shline
Method & AP & AP$_{50}$ & AP$_{75}$ & AP$_s$ & AP$_m$ & AP$_l$ \\
\shline
\multicolumn{7}{c}{\scriptsize{ResNet-50~\cite{resnet}}} \\
RetinaNet \cite{det2}   & 38.7 & 58.0 & 41.5 & 23.3 & 42.3 & 50.3  \\
Faster R-CNN \cite{det2} & 40.2 & 61.0 & 43.8 & 24.2 & 43.5 & 52.0  \\
Cascade R-CNN \cite{det2} & 44.3 &62.2 & 48.0 & 26.6 & 47.7 & 57.7  \\
DETR \cite{detr}   & 42.0 & 62.4 & 44.2 & 20.5 & 45.8 & 61.1  \\
Deformable DETR \cite{deformableDETR}  & 43.8 & 62.6 & 47.7 & 26.4 & 47.1 & 58.0  \\
Sparse R-CNN \cite{sparse-rcnn}   &  \yellowc45.0 & \yellowc63.4 & \yellowc48.2 & \yellowc26.9 & 47.2 & 59.5  \\
\hline
DiffusionDet$^\dagger$ (100@1) & 43.1 & 60.6 & 46.4 & 23.9 & 46.3 & 60.1 \\
DiffusionDet$^\dagger$ (100@2) & 43.0 & 60.4 & 46.6 & 23.9 & 46.1 & 60.1 \\
DiffusionDet$^\dagger$ (100@3) & \orangec45.1 & \orangec63.8 & \orangec48.3 & \orangec27.1 & \orangec48.2 & \yellowc60.7 \\
\diffcell{FlowDet (100@1)}  & \diffcell{43.2} & \diffcell{60.9} & \diffcell{46.7} & \diffcell{24.9} & \diffcell{46.4} & \diffcell{60.0} \\
\diffcell{FlowDet (100@2)}  & \diffcell{44.7} & \diffcell{63.2} & \diffcell{\yellowc48.2} & \diffcell{26.6} & \diffcell{\yellowc47.5} & \diffcell{\orangec60.8} \\
\diffcell{FlowDet (100@3)}  & \diffcell{\redc45.5} & \diffcell{\redc64.1} & \diffcell{\redc49.2} & \diffcell{\redc27.9} & \diffcell{\redc48.4} & \diffcell{\redc61.5} \\
\Xhline{2.\arrayrulewidth}
 
\multicolumn{7}{c}{\scriptsize{ResNet-101~\cite{resnet}}} \\
RetinaNet \cite{det2}   & 40.4 & 60.2 & 43.2 & 24.0 & 44.3 & 52.2 \\
Faster R-CNN \cite{det2}    & 42.0 & 62.5 & 45.9 & 25.2 & 45.6 & 54.6 \\
Cascade R-CNN \cite{mmdetection} & 45.5 & 63.7 & 49.9 & 27.6 & 49.2 & 59.1 \\
DETR \cite{detr}  & 43.5 & 63.8 & 46.4 & 21.9 & 48.0 & 61.8  \\
Sparse R-CNN \cite{sparse-rcnn}  & \redc46.4 & \yellowc64.6 & \yellowc49.5 & \yellowc28.3 & 48.3 & 61.6 \\
\hline
DiffusionDet$^\dagger$ (100@1) & 44.6 & 62.3 & 48.4 & 25.6 & 48.5 & 61.7 \\
DiffusionDet$^\dagger$ (100@2) & 44.4 & 62.1 & 48.3 & 24.8 & 48.1 & 61.9 \\
DiffusionDet$^\dagger$ (100@3) & \orangec46.1 & \orangec64.8 & \orangec49.8 & 27.7 & \redc49.9 & \orangec62.4 \\
\diffcell{FlowDet (100@1)}  & \diffcell{44.0} & \diffcell{62.1} & \diffcell{47.5} & \diffcell{25.7} & \diffcell{47.4} & \diffcell{61.1} \\
\diffcell{FlowDet (100@2)}  & \diffcell{45.7} & \diffcell{64.3} & \diffcell{49.4} & \diffcell{\orangec28.5} & \diffcell{\yellowc48.9} & \diffcell{\yellowc62.2} \\
\diffcell{FlowDet (100@3)}  & \diffcell{\orangec46.1} & \diffcell{\redc65.1} & \diffcell{\redc50.0} & \diffcell{\redc28.7} & \diffcell{\orangec49.5} & \diffcell{\redc62.6} \\
\shline
\end{tabular}
\end{adjustbox}

\caption{\textbf{Comparison of object detector performance on the COCO 2017 validation set}.
Highlighting indicates the {\bf\colorbox[HTML]{FF999A}{best}}, {\it \colorbox[HTML]{FFCC99}{second best}} and {\colorbox[HTML]{FFF8AD}{third best}} result per metric.
\texttt{[$N_{eval}@S$]} denotes the number of prior samples ($N_{eval}$) and number of iteration steps (\texttt{S}).
$^\dagger$We re-train DiffusionDet with the standard 3x schedule inline with other non-DETR models.}
\label{tab:coco}
\vspace{-2mm}
\end{table}

\begin{table}[t]
\tablestyle{4.5pt}{1.1}
\begin{adjustbox}{trim=0pt 0pt 0pt 0pt, clip} 
\begin{tabular}{@{}lcccccc@{}}
\shline
Method & AP & AP$_{50}$ & AP$_{75}$ & AP$_r$ & AP$_c$ & AP$_f$ \\
\shline
\multicolumn{7}{c}{\scriptsize{ResNet-50~\cite{resnet}}} \\
DiffusionDet$^\dagger$ (100@1)  &24.3 &	33.0 &	25.6 &	17.7 &	23.1 &	28.5 \\
DiffusionDet$^\dagger$ (100@2)  &23.7 &	32.6 &	24.8 &	16.1 &	22.5 &	28.3 \\
DiffusionDet$^\dagger$ (100@3)  &\orangec  28.4 &	\orangec 38.9 &	\orangec 29.7 &	\orangec 20.8 &	\redc 27.0 &	\orangec 33.2 \\
\diffcell{FlowDet (100@1)}  & \diffcell{23.4} &	\diffcell{32.3} &	\diffcell{24.5} &	\diffcell{16.6} &	\diffcell{22.1} &	\diffcell{27.9} \\
\diffcell{FlowDet (100@2)}  & \diffcell{\yellowc 27.0} &	\diffcell{\yellowc 37.3} &	\diffcell{\yellowc 28.2} &	\diffcell{\yellowc 20.3} &	\diffcell{\yellowc 25.1} &	\diffcell{\yellowc 32.2} \\
\diffcell{FlowDet (100@3)}  & \diffcell{\redc 28.7} &	\diffcell{\redc 39.7} &	\diffcell{\redc 30.2} &	\diffcell{\redc 22.0} &	\diffcell{\orangec 26.7} &	\diffcell{\redc 33.9} \\
\shline
\end{tabular}
\end{adjustbox}
\caption{\textbf{Comparison of object detector performance on the LVIS v1.0 validation set}.}
\label{tab:lvis}
\vspace{-5mm}
\end{table}

\subsubsection{Discussion}

Overall we present strong evidence validating the improvements provided by CFM's shorter, straighter transport paths, during generative detection. It is worth noting that we have focussed on providing a fair comparison against DiffusionDet, the only previous foray into generative detection. To enable this, we have adopted matching backbones and architectures which are no longer state-of-the-art. However, the concepts implemented in FlowDet are modular and can be adapted
to work with any query set-based detector architecture.
For example we have seen very promising preliminary results using advanced DINO based feature backbones and DETR style architectures. However, in this evaluation we focussed specifically on the differences between the
stochastic SDE formulation of diffusion compared and the linear ODE formulation of CFM. While there
are gains to be found through a greater number of diffusion steps, CFM transport or localisation of
the boxes returns a higher level of performance per integration step. This observation aligns with
the theory behind the two generative formulations. Under compute constraind settings, a more linear path results in less integration error compared to curved stochastic paths.

\subsection{Ablation Study} \label{sec:ablations}

This section details a number of ablations on the prior distribution selection, OT pairing strategy
and ODE solver with runtime analysis. These are areas where CFM provides additional flexibility to FlowDet.
To ease comparison, we fix the backbone to ResNet50, train to a 3x schedule and evaluate with 100 boxes at 1 to 3
integration steps. Additionally, the CFM settings are set to a default of a normal Gaussian prior distribution,
random pairing of prior to ground truth samples and Euler ODE integration when not actively ablated.

\subsubsection{Prior Distribution} \label{sec:abl-priors}

\begin{table}[t]
\tablestyle{5pt}{1.1}

\begin{adjustbox}{trim=0pt 0pt 0pt 0pt, clip} 
\begin{tabular}{@{}lccc@{}}
\shline
Prior Distribution \hspace{2.5cm} & AP & AP$_{50}$ & AP$_{75}$ \\
\shline
\multicolumn{4}{c}{\scriptsize{100 boxes, 1 step}} \\
Gaussian ($\mathcal{D}_{GaussN}$)                 & \redc 43.2 & \redc 60.9 & \redc 46.7 \\
Data-Driven ($\mathcal{D}_{der}$)               & \yellowc42.4 & 59.8 & \yellowc 45.8 \\
Data-Driven size-bucketed ($\mathcal{D}_{der-sb}$) & \yellowc42.4 & \orangec59.9 & 45.4 \\
Data-Dependant ($\mathcal{D}_{dep}$)            & \orangec42.6 & \yellowc60.2 & \orangec45.9 \\
\hline
\multicolumn{4}{c}{\scriptsize{100 boxes, 2 steps}} \\
Gaussian ($\mathcal{D}_{GaussN}$)                  & \redc 44.7 & 63.2 & \orangec48.2 \\
Data-Driven ($\mathcal{D}_{der}$)               & 44.6 & \orangec63.4 & \orangec48.2 \\
Data-Driven size-bucketed ($\mathcal{D}_{der-sb}$) & \redc 44.7 & \redc 64.4 & \redc 49.0 \\
Data-Dependant ($\mathcal{D}_{dep}$)            & \redc 44.7 & \yellowc63.3 & \orangec48.2 \\
\hline
\multicolumn{4}{c}{\scriptsize{100 boxes, 3 steps}} \\
Gaussian ($\mathcal{D}_{GaussN}$)                  & \redc  45.5 & \orangec64.1 & \redc  49.2 \\
Data-Driven ($\mathcal{D}_{der}$)               & 45.0 & 63.4 & 48.2 \\
Data-Driven size-bucketed ($\mathcal{D}_{der-sb}$) & \orangec45.4 & \redc  64.4 & \orangec49.0 \\
Data-Dependant ($\mathcal{D}_{dep}$)            & \yellowc45.2 & \yellowc64.0 & \yellowc48.8 \\
\shline
\end{tabular}
\end{adjustbox}
\caption{\textbf{Prior Distribution selection.}}
\label{tab:abl-priors}
\vspace{-2mm}
\end{table}

Table \ref{tab:abl-priors} quantifies the effect of various prior distributions from which the $x_0$ set of proposal
boxes are sampled. We test a normal Gaussian distribution ($\mathcal{D}_{GaussN}$), data-derived Gaussian parameters (global $\mathcal{D}_{der}$ and
size-bucketed $\mathcal{D}_{der-sb}$) and data-dependant Gaussian parameters ($\mathcal{D}_{dep}$) with the implementations defined in Section
\ref{sec:method-priors}.

In this detection setting, a normal Gaussian performs the best as a prior distribution. We
hypothesise that this is because more compact source distributions increase redundancy between the predicted boxes. In the single-step setting, the resulting loosely packed detection clusters interact poorly with NMS post-processing.
However, it is noteworthy that as the number of inference steps increases, these duplicate
sets tighten, making it easier for NMS to remove duplicates. This closes the gap between
the Gaussian prior and data-derived/dependent priors at higher step counts.

\subsubsection{Optimal Transport Pairing} \label{sec:abl-ot}

\begin{table}[t]
\tablestyle{5pt}{1.1}
\begin{adjustbox}{trim=0pt 0pt 0pt 0pt, clip} 
\begin{tabular}{@{}lccc@{}}
\shline
Cost function \hspace{3cm} & AP & AP$_{50}$ & AP$_{75}$ \\
\shline
\multicolumn{4}{c}{\scriptsize{100 boxes, 1 step}} \\
Random pairing ($\mathcal{M}_{rand}$)  & \redc43.2 & \redc60.9 & \redc46.7 \\
L2 cdist ($\mathcal{M}_{hung-c}$)        & \orangec42.9 & \orangec60.7 & \yellowc46.2 \\
$\ell_1$ + gIoU ($\mathcal{M}_{hung-g}$) & 42.6 & 60.1 & 46.0 \\
InterpIoU ($\mathcal{M}_{hung-i}$)       & \orangec42.9 & \yellowc60.6 & \orangec46.4 \\
\hline
\multicolumn{4}{c}{\scriptsize{100 boxes, 2 step}} \\
Random pairing ($\mathcal{M}_{rand}$) & \redc44.7 & \redc63.2 & \redc48.2 \\
L2 cdist  ($\mathcal{M}_{hung-c}$)       & \yellowc44.3 & \yellowc62.7 & \yellowc47.6 \\
$\ell_1$ + gIoU ($\mathcal{M}_{hung-g}$) & 44.0 & 62.4 & \yellowc47.6 \\
InterpIoU ($\mathcal{M}_{hung-i}$)       & \orangec44.4 & \orangec63.0 & \orangec47.9 \\
\hline
\multicolumn{4}{c}{\scriptsize{100 boxes, 3 step}} \\
Random pairing ($\mathcal{M}_{rand}$)  & \redc45.5 & \redc64.1 & \redc49.2 \\
L2 cdist ($\mathcal{M}_{hung-c}$)       & \yellowc44.8 & \yellowc63.5 & \yellowc48.5 \\
$\ell_1$ + gIoU ($\mathcal{M}_{hung-g}$) & 44.5 & 63.1 & 48.3 \\
InterpIoU ($\mathcal{M}_{hung-i}$)       & \orangec45.0 & \orangec63.9 & \orangec48.6 \\
\shline
\end{tabular}
\end{adjustbox}
\caption{\textbf{OT Pairing Cost Functions.}}
\label{tab:abl-ot}
\vspace{-5mm}
\end{table}

In this ablation, we compare the source-to-target matching strategies described in Section~\ref{sec:matching}.
These include random pairing ($\mathcal{M}_{rand}$) against hungarian matching with three cost functions in
Table \ref{tab:abl-ot}: bounding box centre distance ($\mathcal{M}_{hung-c}$),
bounding box distance + generalised IoU ($\mathcal{M}_{hung-g}$) and InterpIoU ($\mathcal{M}_{hung-i}$).
Interestingly, the results show that any form of coupling, whether primitive or task-aligned, results in worse overall
detector performance. This is likely due to the classification head being starved of easy negative examples
as most interpolants fed into the head are strongly aligned to a ground truth target. In turn, at inference
time, boxes with a high classification scores that do not overlap a targets are
significantly penalised.

Of the remaining matchers, InterpIoU performs the best, with the arrangement of the matchers remaining roughly stable regardless of the number of inference steps.

\begin{table}[t]
\tablestyle{5pt}{1.1}

\begin{adjustbox}{trim=0pt 0pt 0pt 0pt, clip} 
\begin{tabular}{@{}lcccc@{}}
\shline
ODE Solver \hspace{1.75cm}  & AP & AP$_{50}$ & AP$_{75}$ & FPS\\
\shline
\multicolumn{5}{c}{\scriptsize{100 boxes, 1 step}} \\
DiffusionDet (DDIM)  & \orangec 43.1 & \orangec 60.6 & \orangec 46.4 & \redc 73.5 \\
Euler  & \redc 43.2 & \redc 60.9 & \redc 46.7 & \redc 73.5 \\
Heun   & \yellowc 28.6 & \yellowc 40.9 & \yellowc 30.2 & \yellowc 50.0 \\
RK4    & \phantom{0}1.2 & \phantom{0}4.3 & \phantom{0}0.4 & 31.1 \\
\hline
\multicolumn{5}{c}{\scriptsize{100 boxes, 2 steps}} \\
DiffusionDet (DDIM)  & \orangec 43.0 & \orangec 60.4 & \orangec 46.6 & \orangec 49.0 \\
Euler  & \redc 44.7 & \redc 63.2 & \redc 48.2 & \redc 49.5 \\
Heun   & \yellowc 41.8 & \yellowc 59.0 & \yellowc 44.9 & \yellowc 30.3 \\
RK4    & \phantom{0}1.4 & \phantom{0}5.4 & \phantom{0}0.2 & 17.5 \\
\hline
\multicolumn{5}{c}{\scriptsize{100 boxes, 3 steps}} \\
DiffusionDet (DDIM)  & \orangec 45.1 & \orangec 63.8 & \orangec 48.3 & \orangec 37.0 \\
Euler  & \redc 45.5 & \redc 64.1 & \redc 49.2 & \redc 37.3 \\
Heun   & \yellowc 44.2 & \yellowc 62.3 & \yellowc 47.7 & \yellowc 21.9 \\
RK4    & \phantom{0}1.7 & \phantom{0}6.1 & \phantom{0}0.4 & 12.3 \\
\hline
\multicolumn{5}{c}{\scriptsize{100 boxes, adaptive steps}} \\
DOPRI5 1-8 steps, atol=5 & \phantom{0}1.7 & \phantom{0}6.6 & \phantom{0}0.4 & \phantom{0}2.8 \\
DOPRI5 1-16 steps, atol=3 & \phantom{0}1.5 & \phantom{0}5.7 & \phantom{0}0.3 & \phantom{0}1.4 \\
\shline
\end{tabular}
\end{adjustbox}
\caption{\textbf{ODE Solver Methods.} \it{Note: FPS was computed using a single NVIDIA RTX 5090 for inference.}}
\label{tab:abl-solver}
\vspace{-4mm}
\end{table}
\subsubsection{ODE Solver and Runtime Analysis} \label{sec:abl-solver}

Finally, we compare ODE solvers across accuracy and runtime in Table \ref{sec:abl-solver}. 
Euler consistently matches or exceeds the speed of DDIM while achieving greater performance
(e.g., 45.5 vs. 45.1 AP at $\approx$37 FPS, 3 steps). In contrast, higher‑order solvers suffer
at coarser schedules: Heun recovers only by 3 steps and is slower, RK4 \cite{rk-meth} is numerically
brittle with near‑zero AP, and the adaptive DOPRI5 \cite{dopri5} variant fails to reach a good solution
within the step/tolerance caps and inference speed is drastically reduced.

\section{Conclusion}
\label{sec:conclusion}
This paper presents FlowDet, a generalisation of previous diffusion based object detectors to a
broader category of modern generative models. Framing detection as a generative transport task
brings numerous benefits including arbitrarily varying the precision, recall and computational cost
of the predictions at run-time without model retraining.

By adopting a general Conditional Flow Matching formulation, we explore a broad range of source
distributions, matching regimes and ODE solvers. Our results indicate that FlowDet outperforms
DiffusionDet (and other non-generative detection baselines) across a broad range of operating points,
feature backbones and datasets. These broad analyses will provide a solid foundation for future works
in the generative detection field.

{
    \small
    \bibliographystyle{ieeenat_fullname}
    \bibliography{main}
}
\clearpage
\setcounter{page}{1}
\appendix

\section{Formulation of Training Loss} \label{app:train-loss}

We provide a more detailed description of the training supervision, covered in Section
\ref{sec:training-supervision} as follows.

\subsection{$\mathcal{D}_{dep}$ Prior Distribution Supervision}

Given a set of predictions for the data-dependent prior distribution statistics
\((\hat{\mu}, \hat{\sigma})\) as defined in Eqn.~\ref{eqn:data-dep-prior}, we define
the simple Mean-Squared-Error (MSE) loss as
\begin{equation}
    \mathcal{L}_{prior}
    = \lVert \hat{\mu} - \mu \rVert_2^2
    + \lVert \hat{\sigma} - \sigma \rVert_2^2,
\end{equation}
where \((\mu, \sigma)\) are the mean and standard deviation of the ground-truth bounding boxes
computed per batch item.

\subsection{Detection Decoder Supervision}

The outputs from each stage in the detection decoder $(\hat{x}_1, \hat{c})$
are independently supervised against the ground truth set $(x_1, c)$ through
the standard set prediction loss popularised by \cite{detr}. Traditionally,
bipartite matching (via the Hungarian algorithm) between the prediction and
ground-truth sets yields a permutation that minimises a matching cost
\begin{equation} \label{eqn:set-cost-supp}
    \mathcal{C}_{match}
    = \lambda_{cls} \cdot \mathcal{C}_{cls}
    + \lambda_{\ell_1} \cdot \mathcal{C}_{\ell_1}
    + \lambda_{gIoU} \cdot \mathcal{C}_{gIoU}.
\end{equation}
Here, we define $\mathcal{C}_{cls}$ as focal loss \cite{retinanet} between
$\hat{c}$ and $c$, and $\mathcal{C}_{\ell_1}$ and $\mathcal{C}_{gIoU}$ as
the $\ell_1$ and gIoU \cite{gIoU} box costs between $\hat{x}_1$ and $x_1$,
respectively. Additionally, $\lambda_{cls}$, $\lambda_{\ell_1}$ and
$\lambda_{gIoU}$ control the relative weighting of each component.

In our setting there are many more predictions than true objects, so a strict
one-to-one permutation would discard useful supervision signal. Following
\cite{ota-det, yolox, diffdet}, we replace the Hungarian matching with a
one-to-many assignment that selects, for each ground-truth box, the top-$k$
predictions with the lowest cost $\mathcal{C}_{match}$ as positives.
The resulting positive set is used to define the detection loss
\begin{equation} \label{eqn:set-loss}
    \mathcal{L}_{match}
    = \lambda_{cls} \cdot \mathcal{L}_{cls}
    + \lambda_{\ell_1} \cdot \mathcal{L}_{\ell_1}
    + \lambda_{gIoU} \cdot \mathcal{L}_{gIoU},
\end{equation}
with $\mathcal{L}_{cls}$, $\mathcal{L}_{\ell_1}$ and $\mathcal{L}_{gIoU}$
mirroring the components of $\mathcal{C}_{match}$ as in Eqn.~\ref{eqn:set-cost-supp}
but now used as loss terms with the same weights $\lambda_{cls}$, $\lambda_{\ell_1}$ and
$\lambda_{gIoU}$. Predictions that are not selected for any ground-truth
object are ignored in $\mathcal{L}_{match}$. We recompute this top-$k$
assignment independently for every decoder stage and normalise the resulting
loss for each stage by the number of positive matches.

\section{DINOv3 ViT Backbones}

To assess FlowDet with stronger foundation encoders, we further evaluate it
with the DINOv3 ViT-16 backbone \cite{dinov3} in the base (B)
variant. Here, the backbone is kept frozen and we attach a
lightweight simple feature pyramid neck following ViTDet \cite{vitdet}.
In contrast to a standard FPN, this neck constructs a multi-scale hierarchy
directly from the final stride-16 ViT feature map using parallel convolutions
and deconvolutions, without top-down or lateral connections between backbone
stages. This design exploits the strong, single-scale representations learned
by DINOv3 while injecting only minimal detection-specific structure. 
This allows our analysis to focus on the contributions of the FlowDet decoder
when scaling to more powerful backbones, without confounding the findings with
simultaneous innovations in the backbone architecture itself. We
summarise the detection performance of the DINOv3-based backbone configuration
in Table~\ref{tab:dinov3}.

\begin{table}[t]
    \begin{adjustbox}{width=\linewidth}
        \tablestyle{3.7pt}{1.1}
        
\begin{tabular}{@{}lccccccc@{}}
\shline
Backbone & Steps & AP & AP$_{50}$ & AP$_{75}$ & AP$_s$ & AP$_m$ & AP$_l$ \\
\shline
                        & 1 & 43.2 & 60.9 & 46.7 & 24.9 & 46.4 & 60.0 \\
ResNet-50~\cite{resnet} & 2 & 44.7 & 63.2 & 48.2 & 26.6 & 47.5 & 60.8 \\
                        & 3 & 45.5 & 64.1 & 49.2 & \yellowc 27.9 & 48.4 & 61.5 \\

\midrule

                         & 1 & 44.0 & 62.1 & 47.5 & 25.7 & 47.4 & 61.1 \\
ResNet-101~\cite{resnet} & 2 & 45.7 & 64.3 & 49.4 & \orangec 28.5 & 48.9 & 62.2 \\
                         & 3 & \yellowc 46.1 & 65.1 & \yellowc 50.0 & \redc 28.7 & 49.5 & 62.6 \\

\midrule

                                 & 1 & 45.2 & \yellowc 66.1 & 49.1 & 24.1 & \yellowc 50.1 & \yellowc 66.1 \\
DINOv3 ViT-16 Base~\cite{dinov3} & 2 & \orangec 47.1 & \orangec 68.8 & \orangec 51.3 & 26.2 & \orangec 51.6 & \orangec 67.4 \\
                                 & 3 & \redc 47.6 & \redc 69.8 & \redc 51.6 & 27.2 & \redc 52.2 & \redc 67.3 \\
\shline
\end{tabular}

    \end{adjustbox}
    \caption{\textbf{COCO 2017 validation performance of FlowDet with different
    backbones and integration steps.} 
    Highlighting indicates the {\bf\colorbox[HTML]{FF999A}{best}}, {\it \colorbox[HTML]{FFCC99}{second best}} and {\colorbox[HTML]{FFF8AD}{third best}} result per metric.
    All results are obtained with $N_{eval} = 100$
    with the “Steps” column reporting the number of integration steps $S$ used at inference.}
    \label{tab:dinov3}
\end{table}

\begin{figure*}[hp!]
  \centering
  \includegraphics[width=\linewidth]{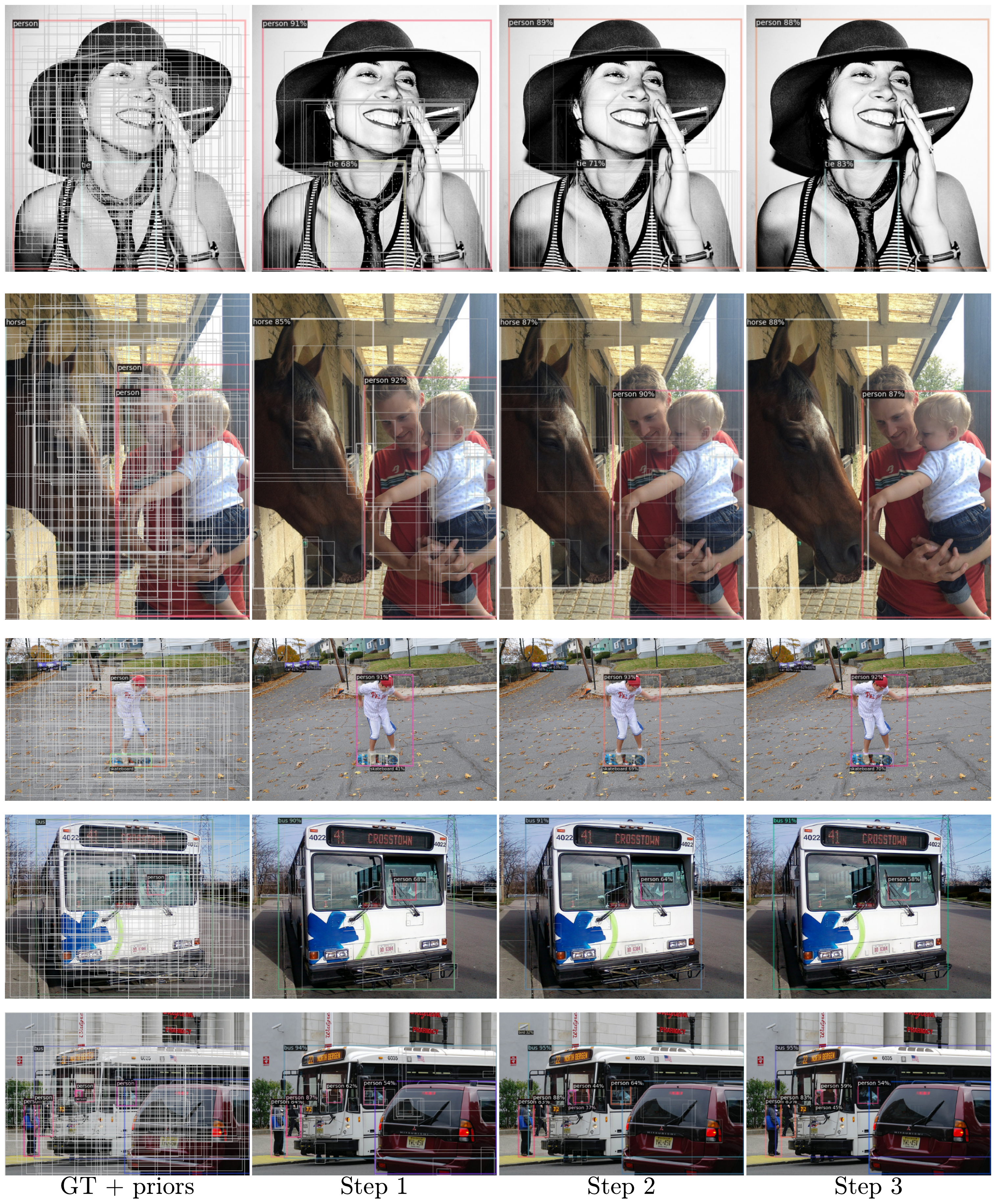}
  \caption{
  \textbf{Qualitative examples of FlowDet on COCO.} For each image, we show
  (from left to right): the input with ground-truth boxes (solid) and initial
  prior samples (faint), followed by the model’s predictions after 1, 2, and 3
  integration steps. All examples use the same backbone, ResNet50, and inference
  settings as our main experiments and $N_{eval} = 100$ per image.}
  \label{fig:qual-ex}
\end{figure*}

Across all settings, using a frozen DINOv3 ViT-16 backbone leads to consistent
improvements over both ResNet baselines (Table \ref{tab:coco})
in these recall-constrained regimes. Impressively, this is achieved without
increasing the compute budgets allocated for training or inference. In particular,
the DINOv3-B configuration achieves 47.6\% AP on COCO, compared to
45.5\% AP with ResNet-50 or 46.1\% AP with ResNet-101. Additionally, the gains
are most pronounced at the AP$_{50}$ threshold with a +5.7\% and +4.7\%
gain over ResNet-50 and ResNet-101 respectively.
These results demonstrate that FlowDet scales favourably with modern backbones such
as DINOv3 without any backbone fine-tuning, and that a simple feature pyramid remains
sufficient when moving from convolutional to ViT encoders.

\clearpage

\section{Qualitative Detection Examples}

To complement the quantitative results of the main paper, we provide qualitative
examples of FlowDet's performance and failure modes in Figure \ref{fig:qual-ex}.
In particular, we illustrate how FlowDet refines detections over multiple flow
steps, allowing boxes to traverse long distances and find objects even with few
initial bounding box candidates. In particular, for a
set of representative COCO images, we show the input with ground-truth boxes,
the initial prior samples, and the model’s predictions after 1, 2, and 3 refinement
steps using the same backbone and evaluation setting as in our main experiments.

As the number of steps increases, boxes are rapidly transported towards tighter,
better-aligned localisations, while spurious and duplicate detections are suppressed,
visually highlighting the effect of the learned flow field on detection quality.
Additionally, we observe a common failure mode of NMS-based detectors in the second row
of examples within Figure \ref{fig:qual-ex}. Here, there are two overlapping ground-truth
boxes of the same class, and valid detection proposals for the smaller instance are
suppressed during NMS, leaving just a single bounding box as a result. It may be possible
to address this using recent refinements to the labelling within COCO \cite{mj-coco} or
through exploring alternative NMS strategies, but we leave this for future work.

\end{document}